\documentclass[letterpaper]{article} % DO NOT CHANGE THIS
\usepackage{aaai2027}  % DO NOT CHANGE THIS
\usepackage[hyphens]{url}  % DO NOT CHANGE THIS
\usepackage{graphicx} % DO NOT CHANGE THIS
\usepackage{natbib}  % DO NOT CHANGE THIS AND DO NOT ADD ANY OPTIONS TO IT
\usepackage{caption} % DO NOT CHANGE THIS AND DO NOT ADD ANY OPTIONS TO IT
\usepackage{amsmath}
\usepackage{amssymb}
\usepackage{booktabs}
\usepackage{multirow}
\usepackage{algorithm}
\usepackage{algorithmic}

\title{VEPO: Mitigating Multilingual Hallucinations in LVLMs via Visual Evidence-guided Preference Optimization}
\author{
    Chengpeng Fu\textsuperscript{\rm 1,2},
    Xiaocheng Feng\textsuperscript{\rm 1,2}\thanks{Corresponding Author.},
    Wenshuai Huo\textsuperscript{\rm 1,2},
    Yichong Huang\textsuperscript{\rm 1},\\
    Baohang Li\textsuperscript{\rm 1},
    Duyu Tang\textsuperscript{\rm 3},
    Ting Liu\textsuperscript{\rm 1,2}
}
\affiliations{
    \textsuperscript{\rm 1}Harbin Institute of Technology, 92 Xidazhi Street, Nangang District, Harbin, Heilongjiang Province, China \\
    
    \textsuperscript{\rm 2}Peng Cheng Laboratory, No.6001 Shahu West Road, Nanshan District, Shenzhen, Guangdong  Province, China\\

    \textsuperscript{\rm 3}Zhongguancun Academy, Beijing, China\\

    \{cpfu, xcfeng, ychuang, wshuo, baohangli, tliu\}@ir.hit.edu.cn
}

\begin{document}

\maketitle

\begin{abstract}
Large vision-language models (LVLMs) achieve strong multimodal performance, yet they remain prone to object hallucination---generating answers that are linguistically fluent but visually ungrounded. This issue is especially severe in low-resource languages, where weaker cross-modal alignment causes models to rely more on language priors than on image evidence. Existing remedies either require costly multilingual annotations or external translation/reward models, or intervene at inference time in ways that can distort the output language. We propose \textbf{VEPO} (Visual Evidence-guided Preference Optimization), a training-time method that transfers reliable visual grounding from English to other languages without external models. VEPO first localizes key visual evidence using English queries via mid-layer cross-modal attention, then constructs preference pairs that share the same correct answer but differ in evidence sufficiency (full image vs.\ key-region masked image). Direct preference optimization on these pairs encourages low-resource inputs to prefer answers supported by complete visual evidence. Experiments on multilingual POPE and MME with Chinese, French, and Hindi show consistent hallucination reduction and perception gains, with the largest improvements on the lowest-resource language. Analyses further confirm that VEPO sharpens question-relevant attention and that evidence-guided masking outperforms random masking.
\end{abstract}

\section{Introduction}

\begin{figure}[t]
\centering
\includegraphics[width=1.0\columnwidth]{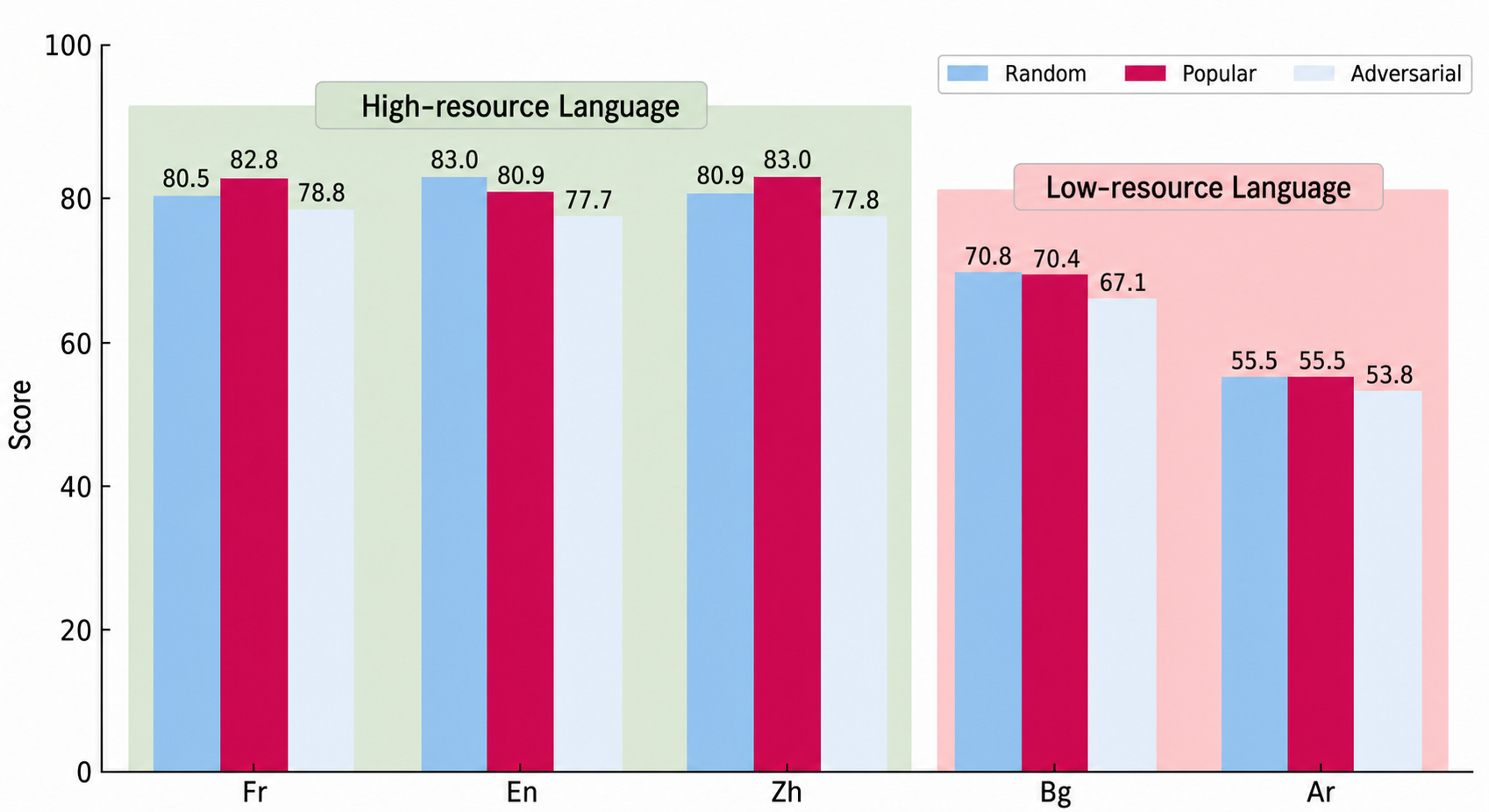}
\caption{Multilingual performance gap of Qwen-VL-Chat on POPE-COCO: non-English, especially low-resource, languages lag behind English.}
\label{fig:multilingual_gap}
\end{figure}

Large vision-language models (LVLMs) have advanced rapidly on visual question answering, image captioning, and multimodal dialogue~\cite{liu2024llava,liu2024improved,bai2023qwenvl-arxiv,openai2023gpt4v}. Despite strong overall capability, they still suffer from \emph{object hallucination}: producing answers that appear plausible yet lack visual support~\cite{li-etal-2023-evaluating,zhou2024analyzing,leng2024mitigating}. Critically, this failure is language-dependent. As shown in Figure~\ref{fig:multilingual_gap}, Qwen-VL-Chat exhibits substantially lower accuracy on low-resource queries than high-resource ones on POPE-COCO, indicating uneven multimodal perception across languages.

We term this phenomenon \emph{multilingual hallucination}: semantically equivalent questions in different languages induce inconsistent evidence retrieval and decision behavior, causing low- resource inputs to hallucinate more frequently. Prior work mitigates the gap mainly through data-centric adaptation. Supervised fine-tuning (SFT) with multilingual visual instructions improves instruction following~\cite{rasheed2025palo}, but high-quality low-resource annotations are expensive and hard to scale. Preference-based methods such as MHR~\cite{qu2026mitigating} construct cross-lingual preference signals via translation or strong multilingual models, thus still relying on external systems. CLAIM~\cite{ye2025claim} instead intervenes in cross-lingual attention at inference time to transfer English visual focus to other languages. While effective for grounding, such test-time intervention can bias generation toward English semantic pathways, yielding visually correct yet linguistically incorrect outputs (e.g., answering a Hindi question in English).

This raises a more fundamental question: can we improve low-resource visual grounding using only mechanisms already present inside the LVLM, without external models or inference-time language distortion? Empirically, for semantically equivalent questions, the image regions that support a correct answer should be language-invariant. In practice, however, attention maps under English and low-resource queries often diverge: English inputs focus on question-relevant objects, whereas low-resource inputs attend more diffusely or to irrelevant regions. This suggests that cross-lingual attention consistency is a useful optimization signal.

We propose \textbf{VEPO} (Visual Evidence-guided Preference Optimization). Rather than preferring a ``correct'' answer over an ``incorrect'' one, VEPO prefers the \emph{same} correct answer under a complete visual input over the same answer under a key-evidence-masked input. Concretely, we (i)~use English queries as probes to identify the top-\(\rho\) image patches most attended during answer generation (default \(\rho{=}0.2\)), (ii)~mask those patches to form rejected images while keeping the original image as the chosen image, and (iii)~apply direct preference optimization (DPO)~\cite{rafailov2024direct,wang-etal-2024-mdpo} on low-resource queries with fixed target answers. Training thus instills a preference for evidence-sufficient grounding while preserving target-language generation, avoiding the language drift of inference-time attention transfer.

Our contributions are threefold:
\begin{itemize}
\item We formulate multilingual hallucination as insufficient visual-evidence reliance under low-resource inputs, and propose VEPO, which constructs evidence-sufficiency preference pairs without external translators or reward models.
\item We introduce an English-attention-guided masking pipeline that localizes key visual evidence from mid-layer attention (layers 11--14 in LLaVA-1.5) and transfers grounding preferences to other languages via training-time DPO.
\item Experiments on multilingual POPE and MME show consistent gains across Chinese, French, and Hindi (e.g., Hindi POPE average F1 \(0.302{\rightarrow}0.644\)); ablations verify the necessity of evidence-guided (vs.\ random) masking and of mixing English data for extremely low-resource languages.
\end{itemize}

\section{Related Work}

\paragraph{Object hallucination in LVLMs.}
Object hallucination has been extensively studied for English-centric LVLMs~\cite{li-etal-2023-evaluating,gunjal2024detecting,zhou2024analyzing}. Mitigation strategies include decoding-time contrast~\cite{leng2024mitigating}, preference/RLHF alignment~\cite{yu2024rlhf,wang-etal-2024-mdpo}, and curated correctional feedback. Most of this literature assumes English evaluation; multilingual degradation remains comparatively underexplored, even though language priors that drive hallucination are often stronger when visual grounding is weak. Closing this gap requires methods that improve evidence use under low-resource prompts, rather than only increasing target-language fluency.

\paragraph{Multilingual multimodal models.}
Multilingual visual instruction tuning improves coverage across languages~\cite{rasheed2025palo}, but residual hallucination persists, especially when visual grounding is weak. Cross-lingual preference learning~\cite{qu2026mitigating} and attention intervention~\cite{ye2025claim} target multilingual hallucination more directly: the former often relies on translation/judging systems, while the latter may distort the output language at inference time. VEPO differs by optimizing \emph{evidence sufficiency} at training time using internal English attention, without external scorers or inference-time intervention.

\paragraph{Preference optimization for multimodal models.}
DPO~\cite{rafailov2024direct} and multimodal variants such as mDPO~\cite{wang-etal-2024-mdpo} improve alignment by contrasting preferred and rejected responses. Standard multimodal DPO typically contrasts different answer texts under a fixed image. VEPO instead contrasts different \emph{image conditions} under a shared answer, explicitly teaching the model when visual evidence is adequate to support that answer. This shift matters for multilingual hallucination: when language priors are strong and visual grounding is weak, answer-level preferences can be satisfied without consulting the image, whereas evidence-sufficiency preferences force the model to notice whether the supporting regions are present.

\section{Method}

\begin{figure*}[t]
\centering
\includegraphics[width=0.93\textwidth]{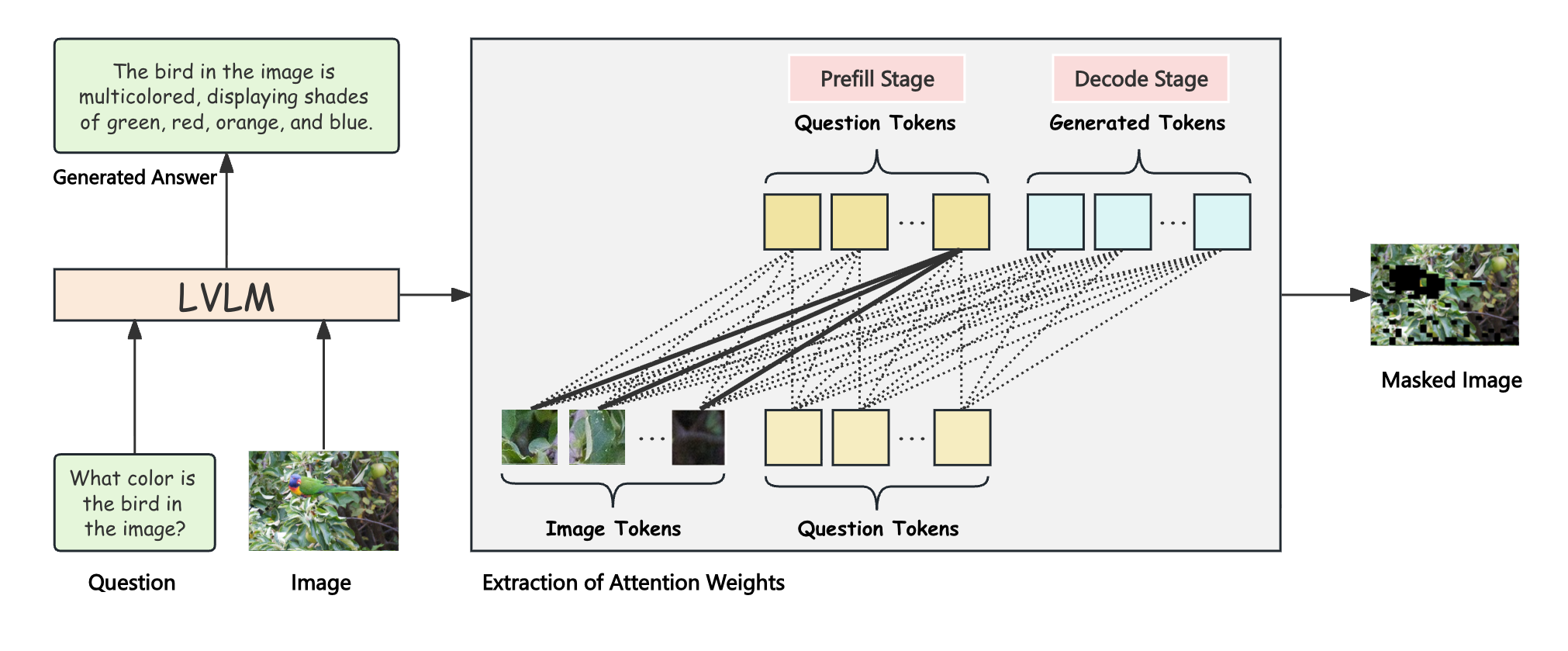}
\caption{Constructing a key-evidence masked image from English attention. Mid-layer attention from the last query token ranks image patches; the top-\(\rho\) patches are blacked out to form the rejected image, while the original image remains the chosen image.}
\label{fig:mask_pipeline}
\end{figure*}

\begin{figure*}[t]
\centering
\includegraphics[width=0.90\textwidth]{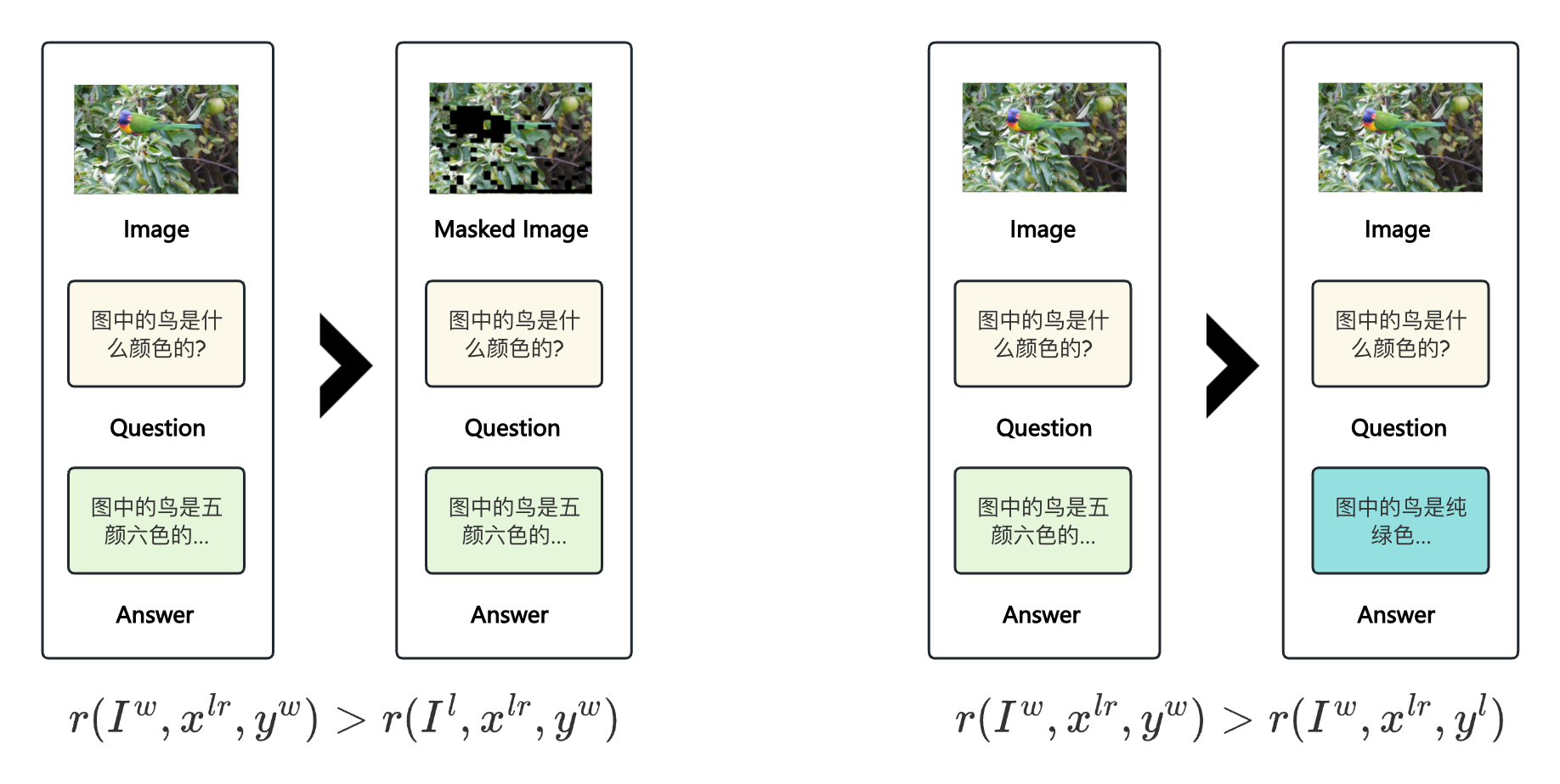}
\caption{Comparison between VEPO (left) and standard multimodal DPO (right). VEPO contrasts evidence-sufficient vs.\ evidence-missing images under the same correct answer; standard DPO contrasts different answers under the same image. The training examples illustrated in the figure are presented in Chinese, and the semantic content of the questions is consistent with that shown in Figure~\ref{fig:mask_pipeline}.}
\label{fig:overview}
\end{figure*}

VEPO consists of three stages: (i)~localizing key visual evidence with an English query, (ii)~constructing an evidence-occluded rejected image, and (iii)~optimizing a low-resource preference objective that favors the same correct answer under complete evidence over incomplete evidence. Figure~\ref{fig:mask_pipeline} illustrates how English mid-layer attention is converted into a key-evidence mask; Figure~\ref{fig:overview} then contrasts the resulting preference construction with standard multimodal DPO. Algorithm~\ref{alg:vepo} summarizes the full pipeline.

\begin{algorithm}[t]
\caption{VEPO Preference Construction and Optimization}
\label{alg:vepo}
\textbf{Input}: Image \(I\), English query \(x^{\mathrm{en}}\), low-resource query \(x^{\mathrm{lr}}\), answer \(y\), mask ratio \(\rho\), layer set \(\mathcal{L}\)\\
\textbf{Output}: Updated policy \(\pi_\theta\)
\begin{algorithmic}[1]
\STATE Run \(\pi_{\mathrm{ref}}\) on \((I,x^{\mathrm{en}})\) and collect mid-layer attentions \(\{A^{(l)}\}_{l\in\mathcal{L}}\)
\STATE Compute evidence scores \(s_i\) from the last query token to each image token
\STATE Select \(\mathcal{S}=\mathrm{TopK}(\{s_i\}, \rho m)\); set \(I^{w}{=}I\), \(I^{l}{=}\mathrm{Mask}(I,\mathcal{S})\)
\STATE Form preference pair \((I^{w},x^{\mathrm{lr}},y) \succ (I^{l},x^{\mathrm{lr}},y)\)
\STATE Update \(\pi_\theta\) by minimizing \(\mathcal{L}_{\mathrm{VEPO}}\) on the constructed pairs
\end{algorithmic}
\end{algorithm}

\subsection{Key Visual Evidence Localization}

Given an image \(I\) and an English query \(x^{\mathrm{en}}=\{t^{\mathrm{en}}_1,\ldots,t^{\mathrm{en}}_n\}\), the LVLM generates an answer autoregressively. We treat English as a high-resource probe language: because English visual grounding is typically stronger after large-scale pretraining and instruction tuning, its cross-modal attention provides a more reliable estimate of which image regions truly support the answer. At layer \(l\), with hidden states \(H^{(l-1)}\),
\begin{equation}
\begin{aligned}
Q^{(l)}&=H^{(l-1)}W_Q^{(l)},\\
K^{(l)}&=H^{(l-1)}W_K^{(l)},\\
V^{(l)}&=H^{(l-1)}W_V^{(l)}.
\end{aligned}
\end{equation}
The attention matrix is \(A^{(l)}=\mathrm{softmax}(Q^{(l)}{K^{(l)}}^\top/\sqrt{d})\).
We use the attention from the last query token \(t_n^{\mathrm{en}}\) to each image token \(v_i\) as an evidence signal. Empirically, this token aggregates the full question semantics after encoding and yields more compact, question-aligned maps than averaging all prompt tokens or using generated tokens (Section~\ref{sec:analysis}). Because shallow layers capture local texture patterns and deep layers are biased toward language generation, we average mid-layer attentions over a layer set \(\mathcal{L}\) (layers 11--14 in LLaVA-1.5-7B):
\begin{equation}
s_i=\frac{1}{|\mathcal{L}|}\sum_{l\in\mathcal{L}}A^{(l)}(t_n^{\mathrm{en}},v_i).
\end{equation}
Higher \(s_i\) indicates stronger support for the English answering process. The resulting ranking of image patches defines the occlusion targets used to build rejected images in the next stage.

\subsection{Key Evidence Occlusion}

Let \(\{s_i\}_{i=1}^{m}\) be scores over \(m\) image tokens. We select the top \(\rho m\) tokens as key evidence,
\begin{equation}
\mathcal{S}=\mathrm{TopK}\bigl(\{s_i\}_{i=1}^{m}, \rho m\bigr),
\end{equation}
and obtain the rejected image by masking the corresponding patches,
\begin{equation}
I^{l}=\mathrm{Mask}(I,\mathcal{S}),\qquad I^{w}=I,
\end{equation}
where \(\rho{=}0.2\) by default. \(\mathrm{Mask}(\cdot)\) zeros the selected patches rather than randomly destroying the image, so \(I^{l}\) retains scene structure and background context while removing question-critical evidence. Intuitively, if English attention for a color question concentrates on a bird's feathers, those patches are blacked out in \(I^{l}\), forcing a hard ``evidence-missing'' negative without inventing a new answer label. This design is crucial: the rejected sample is not an arbitrary corrupted image, but a carefully constructed case where the visual evidence needed for \(y\) is selectively removed.

As shown on the left of Figure~\ref{fig:overview}, VEPO therefore compares two image conditions under a fixed correct answer. In contrast, standard multimodal DPO (right) keeps the image fixed and contrasts a correct answer against an incorrect answer. The former directly supervises visual grounding; the latter mainly supervises answer text preference and can still be satisfied by language priors when the image is under-used. Because the rejected image preserves global scene layout, the preference is not about image realism or captionability, but about whether the patches that English attention marks as decisive remain available to the low-resource query.

\subsection{Evidence-Sufficiency Preference Optimization}

For a low-resource query \(x^{\mathrm{lr}}\) with correct answer \(y\), we form the conditional preference pair
\begin{equation}
(I^{w}, x^{\mathrm{lr}}, y),\qquad (I^{l}, x^{\mathrm{lr}}, y).
\end{equation}
The chosen sample provides complete evidence; the rejected sample shares the same answer text but lacks key visual support. With policy \(\pi_\theta\) and reference model \(\pi_{\mathrm{ref}}\), VEPO minimizes
\begin{equation}
\begin{aligned}
\mathcal{L}_{\mathrm{VEPO}}
&=
-\mathbb{E}_{(I^{w},I^{l},x^{\mathrm{lr}},y)\sim\mathcal{D}}
\Big[
\log \sigma
\Big(
\beta \log \frac{\pi_\theta(y\mid I^{w},x^{\mathrm{lr}})}{\pi_{\mathrm{ref}}(y\mid I^{w},x^{\mathrm{lr}})} \\
&\qquad\qquad
-
\beta \log \frac{\pi_\theta(y\mid I^{l},x^{\mathrm{lr}})}{\pi_{\mathrm{ref}}(y\mid I^{l},x^{\mathrm{lr}})}
\Big)
\Big].
\end{aligned}
\end{equation}
This objective raises the relative likelihood of \(y\) under complete evidence and lowers it under missing evidence, teaching the model to condition answers on visual sufficiency rather than language priors alone. Importantly, because \(y\) remains in the target language for both sides of the pair, VEPO improves grounding without encouraging English-only decoding---a failure mode that inference-time English-attention transfer can introduce.

In practice, VEPO is applied as a lightweight post-training stage: the reference model is frozen, only LoRA adapters are updated, and the preference pairs reuse existing multilingual instruction answers. Consequently, the method does not require new human labels for rejected answers, nor does it depend on an external reward model to score hallucinations. The only language asymmetry is intentional and confined to evidence localization, where English serves as a probe; the optimization itself remains fully target-language.

\section{Experiments}

\subsection{Setup}

\paragraph{Data.}
Training uses the PALO multilingual visual-instruction corpus~\cite{rasheed2025palo} (\({\sim}2.1\)M multilingual visual instructions spanning English, Chinese, French, Spanish, Russian, Japanese, Arabic, Hindi, Bengali, and Urdu). For each selected training sample, we pair the target-language question with its English counterpart on the same image: English is used only to localize evidence, while preference optimization is performed on the target-language question with the original supervised answer \(y\). This design avoids collecting new low-resource annotations and does not require an external translator at training time beyond the already available English parallel questions in PALO. Evaluation uses POPE~\cite{li-etal-2023-evaluating} for object hallucination (F1 on random/popular/adversarial splits over COCO, A-OKVQA, and GQA sources) and MME~\cite{fu2026mme} for perception and cognition (14 subtasks). We report Chinese (Zh), French (Fr), and Hindi (Hi) as representative mid- to low-resource settings.

\paragraph{Models and baselines.}
All methods use LLaVA-1.5-7B~\cite{liu2024improved} as the backbone:
\begin{itemize}
\item \textbf{LLaVA}: pretrained model without further adaptation.
\item \textbf{LLaVA+SFT}: supervised fine-tuning on target-language PALO instructions.
\item \textbf{LLaVA+VEPO}: VEPO applied directly to the pretrained model.
\item \textbf{LLaVA+SFT+VEPO}: SFT followed by VEPO (full pipeline).
\end{itemize}
For Zh/Fr, VEPO pairs are built from target-language data only; for Hi, we mix English pairs at a 1:1 ratio for training stability (ablated in Section~\ref{sec:analysis}).

\paragraph{Implementation.}
We fine-tune with LoRA (rank 128, \(\alpha{=}256\)) for one epoch, learning rate \(2{\times}10^{-5}\), batch size 8, DPO temperature \(\beta{=}0.1\), and mask ratio \(\rho{=}0.2\). Mid-layer attention uses layers 11--14 and the last query token, validated in Section~\ref{sec:analysis}. The reference model \(\pi_{\mathrm{ref}}\) is frozen as the initialization of \(\pi_\theta\). All multilingual POPE/MME prompts follow the official English templates translated into the target language with the same yes/no answer format. Unless otherwise noted, we report a single run with the above hyperparameters; qualitative attention analyses use the same checkpoints as the quantitative tables.

\subsection{Main Results on POPE}

\begin{table*}[t]
\centering
\setlength{\tabcolsep}{2.2pt}
\footnotesize
\begin{tabular}{@{}llcccccccccc@{}}
\toprule
\multirow{2}{*}{Lang.} & \multirow{2}{*}{Method} & \multicolumn{3}{c}{COCO} & \multicolumn{3}{c}{AOKVQA} & \multicolumn{3}{c}{GQA} & \multirow{2}{*}{Avg.} \\
\cmidrule(lr){3-5} \cmidrule(lr){6-8} \cmidrule(lr){9-11}
& & Random & Popular & Adversarial & Random & Popular & Adversarial & Random & Popular & Adversarial & \\
\midrule
\multicolumn{12}{c}{Original Models} \\
\midrule
\multirow{2}{*}{Zh}
& LLaVA-1.5 & 0.786 & 0.764 & 0.731 & 0.745 & 0.711 & 0.693 & 0.752 & 0.702 & 0.692 & 0.731 \\
& +VEPO & \textbf{0.818} & \textbf{0.815} & \textbf{0.766} & \textbf{0.783} & \textbf{0.764} & \textbf{0.720} & \textbf{0.787} & \textbf{0.731} & \textbf{0.707} & \textbf{0.766} \\
\midrule
\multirow{2}{*}{Fr}
& LLaVA-1.5 & 0.765 & 0.798 & 0.743 & 0.744 & 0.750 & 0.710 & 0.744 & 0.722 & 0.706 & 0.742 \\
& +VEPO & \textbf{0.822} & \textbf{0.858} & \textbf{0.784} & \textbf{0.798} & \textbf{0.814} & \textbf{0.745} & \textbf{0.798} & \textbf{0.783} & \textbf{0.743} & \textbf{0.794} \\
\midrule
\multirow{2}{*}{Hi}
& LLaVA-1.5 & 0.345 & 0.332 & 0.337 & 0.297 & 0.302 & 0.298 & 0.267 & 0.273 & 0.270 & 0.302 \\
& +VEPO & \textbf{0.632} & \textbf{0.669} & \textbf{0.661} & \textbf{0.631} & \textbf{0.639} & \textbf{0.637} & \textbf{0.635} & \textbf{0.656} & \textbf{0.633} & \textbf{0.644} \\
\midrule
\multicolumn{12}{c}{Models after SFT} \\
\midrule
\multirow{2}{*}{Zh}
& +SFT & 0.823 & 0.822 & 0.775 & 0.765 & 0.755 & 0.715 & 0.769 & 0.747 & 0.718 & 0.765 \\
& +SFT+VEPO & \textbf{0.845} & \textbf{0.839} & \textbf{0.782} & \textbf{0.809} & \textbf{0.790} & \textbf{0.737} & \textbf{0.811} & \textbf{0.758} & \textbf{0.726} & \textbf{0.789} \\
\midrule
\multirow{2}{*}{Fr}
& +SFT & 0.768 & 0.730 & 0.721 & 0.736 & 0.716 & 0.702 & 0.737 & 0.716 & 0.705 & 0.726 \\
& +SFT+VEPO & \textbf{0.797} & \textbf{0.834} & \textbf{0.764} & \textbf{0.773} & \textbf{0.783} & \textbf{0.726} & \textbf{0.770} & \textbf{0.746} & \textbf{0.719} & \textbf{0.768} \\
\bottomrule
\end{tabular}
\caption{POPE F1 of VEPO on original and SFT models across Chinese (Zh), French (Fr), and Hindi (Hi). Best results within each language block are in bold.}
\label{tab:pope}
\end{table*}

Table~\ref{tab:pope} reports F1 scores. On the pretrained backbone, VEPO improves average F1 for Zh (\(0.731{\rightarrow}0.766\), \({+}3.5\)), Fr (\(0.742{\rightarrow}0.794\), \({+}5.2\)), and Hi (\(0.302{\rightarrow}0.644\), \({+}34.2\)). Gains are largest for Hindi, consistent with transferring English visual localization to languages with weaker native grounding. Improvements hold across COCO/A-OKVQA/GQA and across random/popular/adversarial splits. Notably, adversarial F1 also rises (e.g., Zh COCO \(0.731{\rightarrow}0.766\); Hi COCO \(0.337{\rightarrow}0.661\)), indicating that VEPO reduces reliance on co-occurrence priors rather than only correcting easy random negatives. In other words, even when distractor objects frequently co-occur with queried categories, the model becomes less likely to affirm their presence without sufficient visual support.

After SFT, VEPO still helps: Zh \(0.765{\rightarrow}0.789\) and Fr \(0.726{\rightarrow}0.768\). Thus SFT and VEPO are complementary---SFT improves instruction following and target-language fluency, while VEPO strengthens evidence-based answering and further suppresses object hallucination. This complementarity is practically useful: one can first adapt an LVLM to a target language with standard supervised data, then apply VEPO as a lightweight second stage to specifically harden visual grounding.

We further ablate two design choices on Chinese and Hindi POPE. Table~\ref{tab:random_mask} contrasts evidence-guided masking with random masking at the same occlusion ratio, and Table~\ref{tab:hindi_en} studies whether mixing English preference pairs stabilizes Hindi training.

\begin{table*}[t]
\centering
\setlength{\tabcolsep}{4.5pt}
\small
\begin{tabular}{lcccccccccc}
\toprule
Method & \multicolumn{3}{c}{COCO} & \multicolumn{3}{c}{AOKVQA} & \multicolumn{3}{c}{GQA} & Avg. \\
\cmidrule(lr){2-4} \cmidrule(lr){5-7} \cmidrule(lr){8-10}
& rand. & pop. & adv. & rand. & pop. & adv. & rand. & pop. & adv. & \\
\midrule
LLaVA-1.5 & 0.786 & 0.764 & 0.731 & 0.745 & 0.711 & 0.693 & 0.752 & 0.702 & 0.692 & 0.731 \\
+Random Mask & 0.793 & 0.789 & 0.743 & 0.753 & 0.731 & 0.706 & 0.767 & 0.719 & 0.704 & 0.745 \\
+VEPO & \textbf{0.818} & \textbf{0.815} & \textbf{0.766} & \textbf{0.783} & \textbf{0.764} & \textbf{0.720} & \textbf{0.787} & \textbf{0.731} & \textbf{0.707} & \textbf{0.766} \\
\bottomrule
\end{tabular}
\caption{Chinese POPE: evidence-guided masking vs.\ random masking under the same occlusion ratio \(\rho{=}0.2\).}
\label{tab:random_mask}
\end{table*}

\begin{table}[t]
\centering
\setlength{\tabcolsep}{6pt}
\small
\begin{tabular}{lcccc}
\toprule
Method & rand. & pop. & adv. & Avg. \\
\midrule
LLaVA-1.5 & 0.345 & 0.332 & 0.337 & 0.338 \\
VEPO (Hi only) & 0.309 & 0.499 & 0.409 & 0.406 \\
VEPO (Hi+En) & \textbf{0.632} & \textbf{0.669} & \textbf{0.661} & \textbf{0.654} \\
\midrule
SFT & 0.706 & 0.711 & 0.709 & 0.709 \\
SFT+VEPO (Hi only) & 0.686 & 0.724 & 0.689 & 0.700 \\
SFT+VEPO (Hi+En) & \textbf{0.772} & \textbf{0.783} & \textbf{0.752} & \textbf{0.769} \\
\bottomrule
\end{tabular}
\caption{Hindi COCO-POPE: mixing English preference data stabilizes VEPO.}
\label{tab:hindi_en}
\end{table}

\subsection{Results on MME}

\begin{figure}[t]
\centering
\includegraphics[width=\columnwidth]{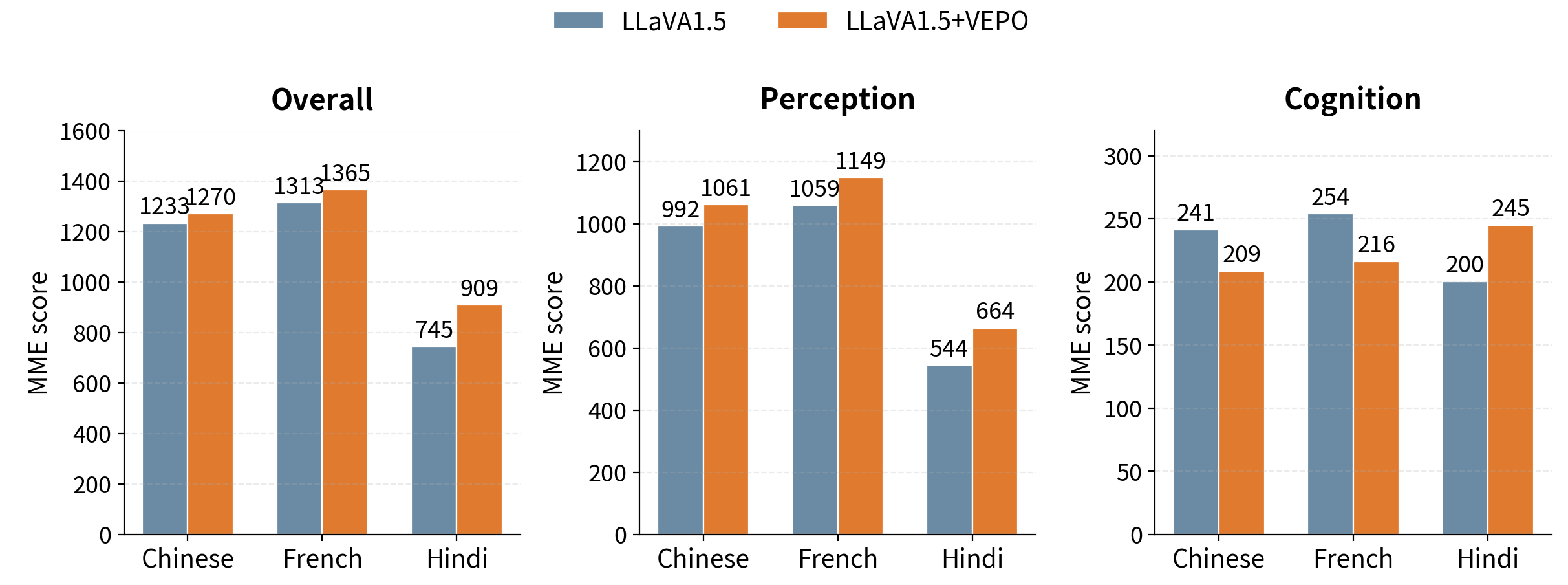}
\caption{Overall MME scores before and after VEPO across languages.}
\label{fig:mme_overall}
\end{figure}

\begin{figure}[t]
\centering
\includegraphics[width=\columnwidth]{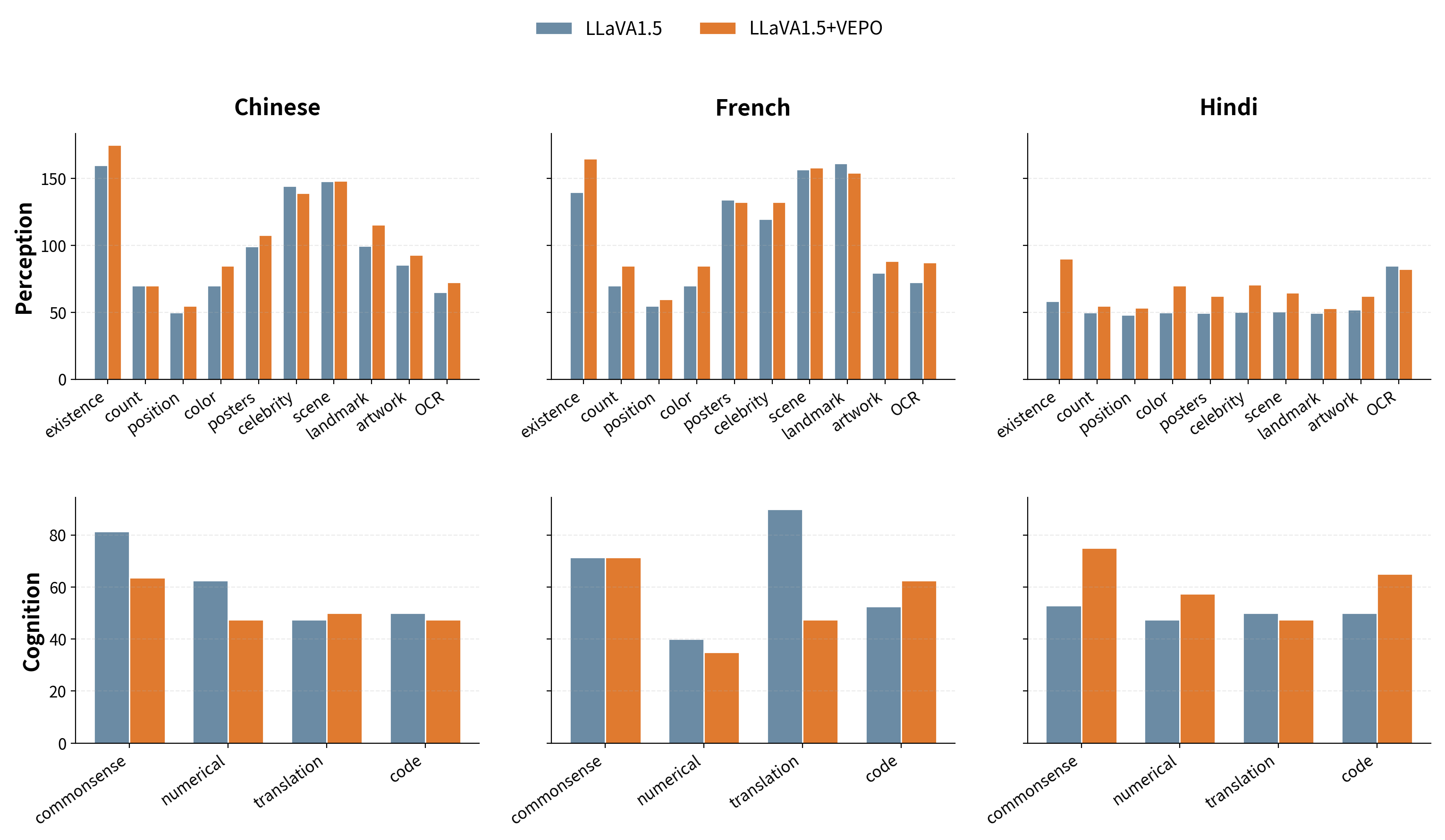}
\caption{Fine-grained MME results: gains are concentrated on perception-oriented subtasks.}
\label{fig:mme_detail}
\end{figure}

Figures~\ref{fig:mme_overall} and~\ref{fig:mme_detail} summarize MME scores. Overall scores rise for Zh (\(1233{\rightarrow}1270\)), Fr (\(1313{\rightarrow}1365\)), and Hi (\(745{\rightarrow}909\)). Perception accounts for most of the gain: Zh \(992{\rightarrow}1061\), Fr \(1059{\rightarrow}1149\), Hi \(544{\rightarrow}664\), with clear improvements on existence, color, position, poster, artwork, and OCR---tasks that require precise local visual evidence. Cognition changes are mixed: Zh/Fr show mild drops on some reasoning subtasks (e.g., commonsense and numerical calculation), while Hi improves (\(200{\rightarrow}245\)). Importantly, total scores still increase in all languages. For Hindi, both perception and cognition rise, suggesting that stronger visual evidence can also stabilize downstream reasoning when language ability itself is weak. Overall, MME corroborates POPE: VEPO mainly improves visually grounded perception, with the largest absolute benefit in the lowest-resource setting.

\section{Analysis}
\label{sec:analysis}

\begin{figure}[t]
\centering
\includegraphics[width=\columnwidth,height=1.0\textheight,keepaspectratio]{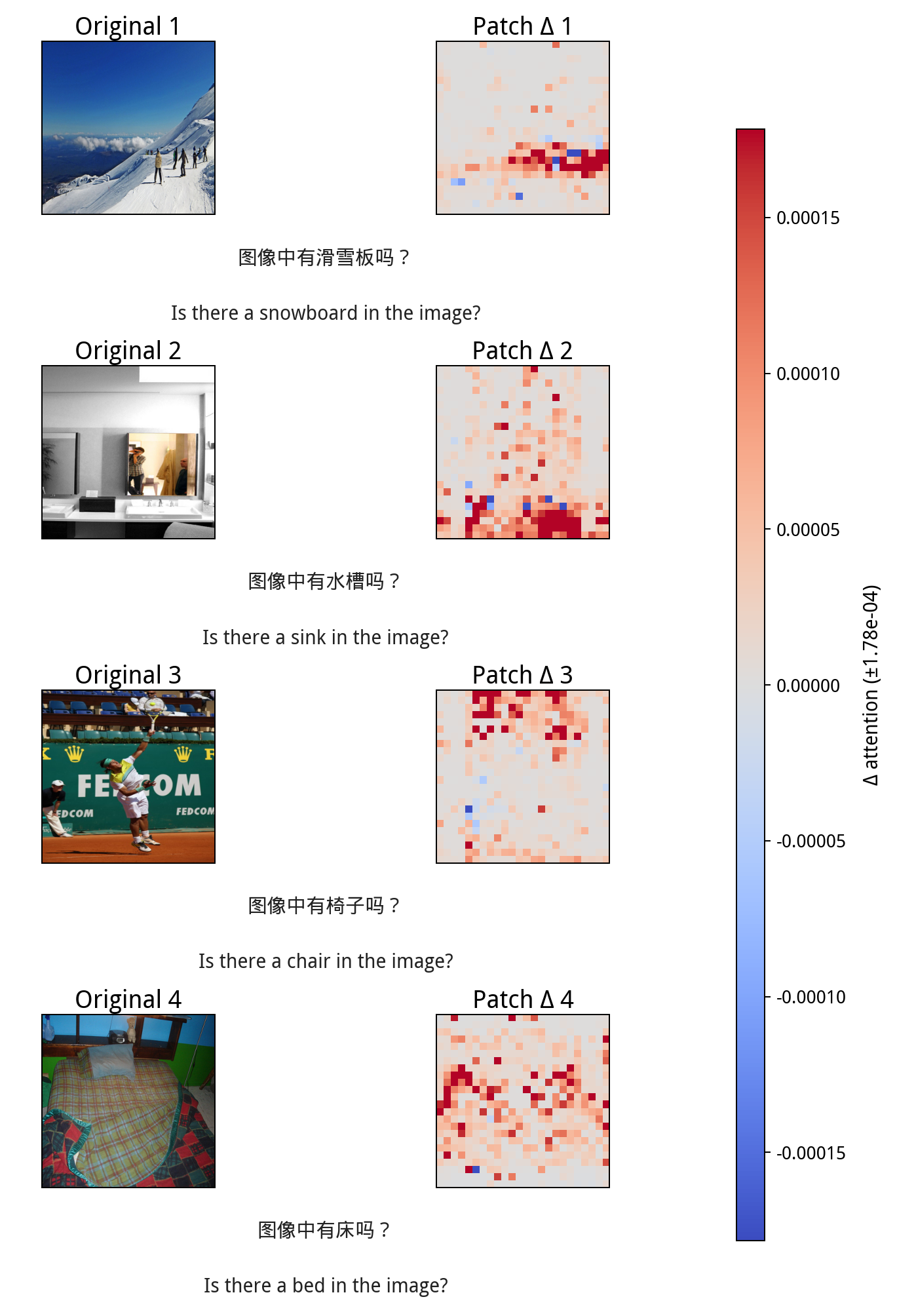}
\caption{Attention change after VEPO: red/blue indicate increased/decreased attention on image patches.}
\label{fig:attn_change}
\end{figure}

\paragraph{Attention redistribution.}
To verify that VEPO redirects attention toward question-relevant evidence, we visualize patch-level attention changes before vs.\ after training (Figure~\ref{fig:attn_change}; red: increase; blue: decrease). VEPO does not uniformly strengthen or weaken visual attention; instead, it exhibits a clear question-conditioned redistribution: gains concentrate on regions needed to answer the query, while attention on irrelevant areas decreases.

Concrete cases match this pattern. For ``Is there a snowboard?'', attention rises near the person and ski-slope activity regions rather than sky or distant background, indicating a shift from scene-level priors to local evidence supporting the queried object. For ``Is there a sink?'', gains localize around the washbasin area in the lower image, while furniture and person regions are relatively suppressed. For ``Is there a chair?'', attention increases on audience seating rather than the salient foreground athlete, showing that VEPO can retrieve background objects that are diagnostic but visually less dominant. For ``Is there a bed?'', gains cover the bed surface and nearby structure, again aligning with the queried category.

Overall, VEPO is semantically selective: it amplifies regions that support the correct answer, not necessarily the most salient foreground. Regions that previously attracted attention but are weakly related to the question are suppressed. This suggests that VEPO strengthens an evidence-based answering preference and transfers a more English-like visual localization pattern to low-resource inputs, rather than only changing final yes/no logits.

\begin{figure}[t]
\centering
\includegraphics[width=\columnwidth,height=1.0\textheight,keepaspectratio]{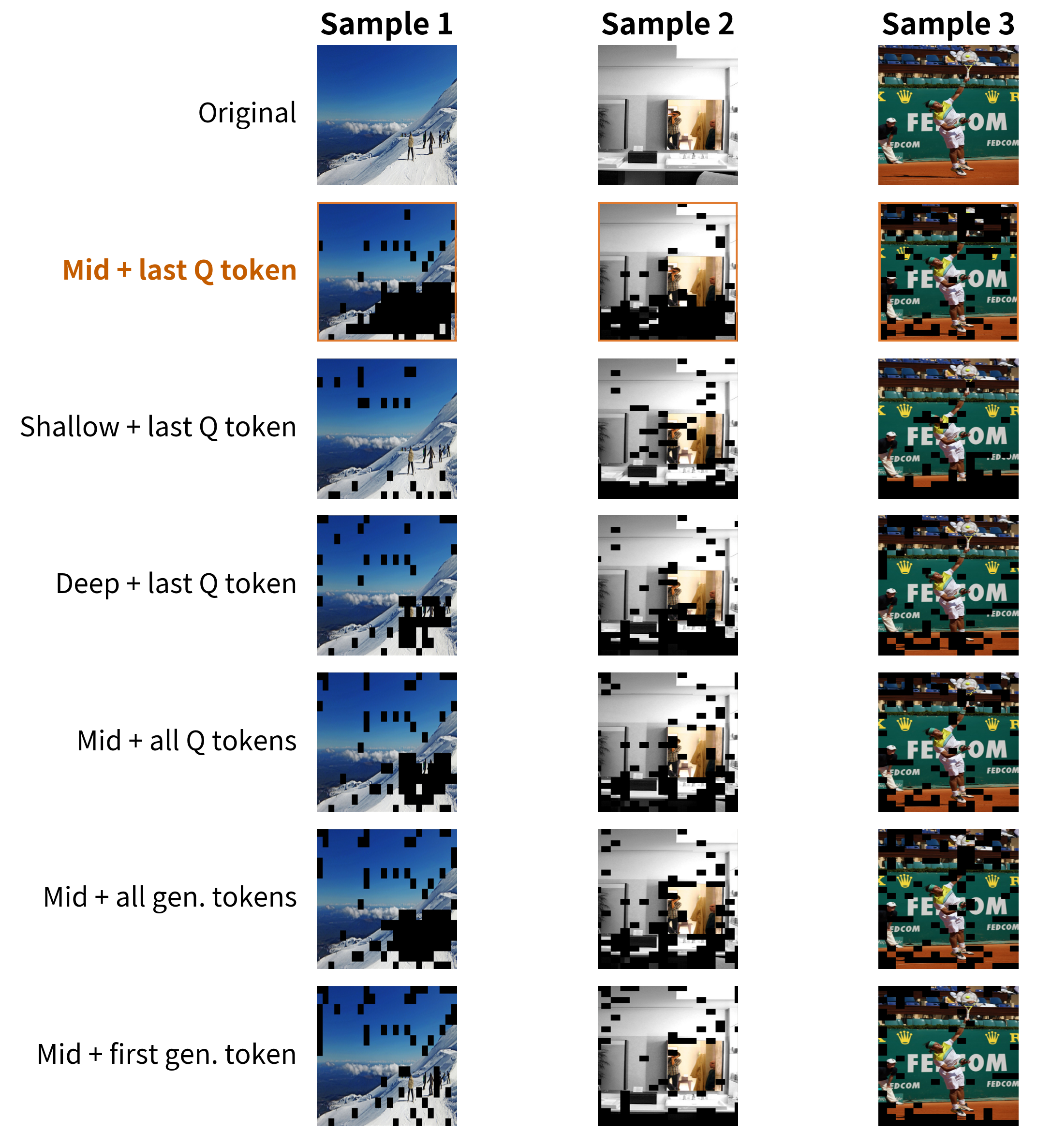}
\caption{Effect of layer depth and token selection on English-attention masks used for VEPO.}
\label{fig:layer_token}
\end{figure}

\paragraph{Layer and token choice for masking.}
When constructing English-guided masks, we black out patches ranked by mid-layer attention from the last query token. To justify this choice, we compare alternative constructions along two axes (Figure~\ref{fig:layer_token}): (i)~layer depth---shallow, mid, and deep attentions (and mid-layer averages); and (ii)~token choice---last query token, all prompt tokens, first generated token, and all generated tokens (averaged when multiple tokens are used). The goal is to identify which language-side anchor and which depth best recover the visual evidence needed for the answer.

Layer comparisons show a clear progression. Shallow masks are sparse and fragmented: they often hit texture edges (sky/snow borders, room bottoms, scattered background) and only partially cover foreground objects, reflecting local appearance patterns rather than stable semantic localization. Mid-layer masks are more compact and question-aligned: they cover activity regions on the slope, person--mirror interactions indoors, and athlete/court regions in sports scenes, while reducing random background responses. This indicates that mid layers are where vision--language alignment becomes effective. Deep masks retain some object focus but shrink, shift, or collapse into blocky clusters, consistent with representations being compressed toward answer generation; intervening with deep attention therefore risks decoding bias rather than reliable evidence localization. These observations motivate averaging layers 11--14 in LLaVA-1.5.

Token comparisons likewise favor the last query token. After full question encoding, this token aggregates the main query semantics and yields the most compact, question-driven masks. Averaging all prompt tokens mixes content words with function words and instruction templates, producing larger, more fragmented masks that spill into background. Attentions from generated tokens are biased by decoding: they partly serve answer organization rather than evidence retrieval, and later tokens further depend on already generated text. First-generated-token masks are more focused than full-generation averages but remain less stable than last-query-token masks. Thus, for both interpretability and intervention quality, the last query token is the preferred language-side anchor.

These analyses also clarify \emph{why} VEPO helps beyond output-level preference shifts. Before training, attention under complex scenes often spreads to weakly related candidates, occluders, or background clutter, encouraging false positives or over-interpretation of local cues. After VEPO, attention concentrates on entity-, attribute-, or relation-relevant regions; when those regions are masked, the model becomes more sensitive to missing evidence and less willing to affirm presence without support. This matches the training signal: contrasting full vs.\ evidence-occluded images teaches the model to separate evidence-sufficient from evidence-missing conditions, improving intermediate evidence selection rather than only final answers.

\paragraph{Evidence-guided vs.\ random masking.}
Table~\ref{tab:random_mask} replaces evidence-guided occlusion with random occlusion at the same ratio \(\rho{=}0.2\). Random masking yields only a modest gain (avg.\ \(0.731{\rightarrow}0.745\)), whereas evidence-guided VEPO reaches \(0.766\), with consistent gaps across COCO/A-OKVQA/GQA and across random/popular/adversarial splits. Random masks frequently remove background or unrelated objects and therefore fail to create informative ``evidence-missing'' negatives: the occluded image remains nearly equally answerable, so the preference pair provides a weak learning signal about when confidence should drop. Evidence-guided masks preferentially remove question-critical patches, producing harder negatives that force the model to link answers to specific visual support. Hence VEPO's gains are not from occlusion augmentation alone, but from targeted destruction of decisive evidence; both the preference objective and the region-selection strategy are necessary.

\paragraph{Mixing English data for low-resource languages.}
Table~\ref{tab:hindi_en} studies whether VEPO should mix English preference pairs for Hindi. Target-only Hindi VEPO is unstable: without SFT, COCO-POPE random F1 drops from \(0.345\) to \(0.309\) and the average is only \(0.406\); after SFT, Hi-only VEPO (\(0.700\)) even falls below SFT alone (\(0.709\)). Mixing English pairs at 1:1 raises averages to \(0.654\) (no SFT) and \(0.769\) (with SFT), with balanced gains on random/popular/adversarial splits. Qualitatively, Hindi-only optimization can collapse to overly conservative ``no'' answers or unstable yes/no decisions, while Hi+En training yields consistent improvements.

We attribute this to data scale and distribution sensitivity. The backbone has limited Hindi exposure, so optimizing solely on scarce Hindi preference pairs can over-adapt parameters toward a small target-language distribution and damage previously stable vision--language alignments. English pairs supply denser multimodal gradients and act as cross-lingual regularization that constrains update magnitude. For higher-resource Zh/Fr, target-only VEPO already suffices; English mixing is therefore reserved for extremely low-resource languages in our main experiments. A practical recipe follows: start with target-only pairs, and introduce English pairs when validation hallucination metrics become unstable or degrade relative to the SFT baseline.

\section{Conclusion}

We presented VEPO, a visual evidence-guided preference optimization method for mitigating multilingual hallucinations in LVLMs. By localizing key evidence with English mid-layer attention and optimizing preferences between full and evidence-masked images under a shared target-language answer, VEPO strengthens low-resource visual grounding without external models or inference-time language distortion. Experiments on POPE and MME demonstrate consistent gains---especially for low-resource Hindi, where average POPE F1 rises from \(0.302\) to \(0.644\)---and analyses confirm improved attention focus, the superiority of evidence-guided over random masking, and the need to mix English preference data for extremely low-resource languages. Limitations include English-centric evidence transfer and possible retuning of mid-layer indices for other LVLM architectures. 
% Future work includes joint multilingual evidence alignment, broader multilingual VQA evaluation, and distilling VEPO-style preferences into lighter inference-time adapters.

% \clearpage
\bibliography{aaai2027}

@inproceedings{liu2024llava,
  title     = {Visual Instruction Tuning},
  author    = {Liu, Haotian and Li, Chunyuan and Wu, Qingyang and Lee, Yong Jae},
  booktitle = {Advances in Neural Information Processing Systems},
  volume    = {36},
  year      = {2023}
}

@inproceedings{liu2024improved,
  title     = {Improved Baselines with Visual Instruction Tuning},
  author    = {Liu, Haotian and Li, Chunyuan and Li, Yuheng and Lee, Yong Jae},
  booktitle = {Proceedings of the IEEE/CVF Conference on Computer Vision and Pattern Recognition},
  pages     = {26296--26306},
  year      = {2024}
}

@misc{bai2023qwenvl-arxiv,
  title         = {{Qwen-VL}: A Versatile Vision-Language Model for Understanding, Localization, Text Reading, and Beyond},
  author        = {Bai, Jinze and Bai, Shuai and Yang, Shusheng and Wang, Shijie and Tan, Sinan and Wang, Peng and Lin, Junyang and Zhou, Chang and Zhou, Jingren},
  year          = {2023},
  eprint        = {2308.12966},
  archivePrefix = {arXiv}
}

@misc{openai2023gpt4v,
  title  = {{GPT-4V(ision)} System Card},
  author = {{OpenAI}},
  year   = {2023},
  note   = {Technical Report}
}

@inproceedings{li-etal-2023-evaluating,
  title     = {Evaluating Object Hallucination in Large Vision-Language Models},
  author    = {Li, Yifan and Du, Yifan and Zhou, Kun and Wang, Jinpeng and Zhao, Xin and Wen, Ji-Rong},
  booktitle = {Proceedings of the 2023 Conference on Empirical Methods in Natural Language Processing},
  pages     = {292--305},
  year      = {2023},
  publisher = {Association for Computational Linguistics}
}

@inproceedings{zhou2024analyzing,
  title     = {Analyzing and Mitigating Object Hallucination in Large Vision-Language Models},
  author    = {Zhou, Yiyang and Cui, Chenliang and Yoon, Jaehong and Zhang, Linjun and Deng, Zhun and Finn, Chelsea and Bansal, Mohit and Yao, Huaxiu},
  booktitle = {The Twelfth International Conference on Learning Representations},
  year      = {2024}
}

@inproceedings{leng2024mitigating,
  title     = {Mitigating Object Hallucinations in Large Vision-Language Models through Visual Contrastive Decoding},
  author    = {Leng, Sicong and Zhang, Hang and Chen, Guanzheng and Li, Xin and Lu, Shijian and Liu, Chunyan and Zhang, Henghui},
  booktitle = {Proceedings of the IEEE/CVF Conference on Computer Vision and Pattern Recognition},
  pages     = {13872--13882},
  year      = {2024}
}

@inproceedings{gunjal2024detecting,
  title     = {Detecting and Preventing Hallucinations in Large Vision Language Models},
  author    = {Gunjal, Anisha and Yin, Jihan and Bas, Erhan},
  booktitle = {Proceedings of the AAAI Conference on Artificial Intelligence},
  volume    = {38},
  number    = {16},
  pages     = {18135--18143},
  year      = {2024}
}

@inproceedings{yu2024rlhf,
  title     = {{RLHF-V}: Towards Trustworthy {MLLMs} via Behavior Alignment from Fine-Grained Correctional Human Feedback},
  author    = {Yu, Tianyu and Yao, Yuan and Zhang, Haoye and He, Taiwen and Han, Yifeng and Cui, Ganqu and Hu, Jinyi and Liu, Zhiyuan and Zheng, Hai-Tao and Sun, Maosong and others},
  booktitle = {Proceedings of the IEEE/CVF Conference on Computer Vision and Pattern Recognition},
  pages     = {13875--13885},
  year      = {2024}
}

@article{rafailov2024direct,
  title   = {Direct Preference Optimization: Your Language Model Is Secretly a Reward Model},
  author  = {Rafailov, Rafael and Sharma, Archit and Mitchell, Eric and Manning, Christopher D and Ermon, Stefano and Finn, Chelsea},
  journal = {Advances in Neural Information Processing Systems},
  volume  = {36},
  year    = {2024}
}

@inproceedings{wang-etal-2024-mdpo,
  title     = {{mDPO}: Conditional Preference Optimization for Multimodal Large Language Models},
  author    = {Wang, Fei and Zhou, Wenxuan and Huang, James Y. and Xu, Nan and Zhang, Sheng and Poon, Hoifung and Chen, Muhao},
  booktitle = {Proceedings of the 2024 Conference on Empirical Methods in Natural Language Processing},
  pages     = {8078--8088},
  year      = {2024},
  publisher = {Association for Computational Linguistics}
}

@inproceedings{rasheed2025palo,
  title     = {{PALO}: A Large Multilingual Multimodal Language Model},
  author    = {Rasheed, Hanoona and Maaz, Muhammad and Shaker, Abdelrahman and Khan, Salman and Cholakal, Hisham and Anwer, Rao M and Baldwin, Tim and Felsberg, Michael and Khan, Fahad S},
  booktitle = {Proceedings of the IEEE/CVF Winter Conference on Applications of Computer Vision},
  year      = {2025}
}

@article{qu2026mitigating,
  title     = {Mitigating Multilingual Hallucination in Large Vision-Language Models},
  author    = {Qu, Xiaoye and Song, Mingyang and Wei, Wei and Liu, Daizong and Dong, Jianfeng and Cheng, Yu},
  journal   = {ACM Transactions on Multimedia Computing, Communications and Applications},
  volume    = {22},
  number    = {5},
  pages     = {1--24},
  year      = {2026},
  publisher = {ACM}
}

@inproceedings{ye2025claim,
  title     = {{CLAIM}: Mitigating Multilingual Object Hallucination in Large Vision-Language Models with Cross-Lingual Attention Intervention},
  author    = {Ye, Zekai and Li, Qiming and Feng, Xiaocheng and Qin, Libo and Huang, Yichong and Li, Baohang and Jiang, Kui and Xiang, Yang and Zhang, Zhirui and Lu, Yunfei and others},
  booktitle = {Proceedings of the 63rd Annual Meeting of the Association for Computational Linguistics (Volume 1: Long Papers)},
  pages     = {13080--13094},
  year      = {2025}
}

@article{fu2026mme,
  title   = {{MME}: A Comprehensive Evaluation Benchmark for Multimodal Large Language Models},
  author  = {Fu, Chaoyou and Chen, Peixian and Shen, Yunhang and Qin, Yulei and Zhang, Mengdan and Lin, Xu and Yang, Jinrui and Zheng, Xiawu and Li, Ke and Sun, Xing and others},
  journal = {Advances in Neural Information Processing Systems},
  volume  = {38},
  year    = {2026}
}

\end{document}